\documentclass{article}

     \usepackage[preprint,nonatbib]{neurips_2024}

\usepackage[utf8]{inputenc} 
\usepackage[T1]{fontenc}    
\usepackage{hyperref}       
\usepackage{url}            
\usepackage{booktabs}       
\usepackage{amsfonts}       
\usepackage{nicefrac}       
\usepackage{microtype}      
\usepackage{xcolor}         

\usepackage{graphicx}
\usepackage{amsmath}
\usepackage{multirow}

\title{GFlowNet Fine-tuning for Diverse Correct Solutions in Mathematical Reasoning Tasks}

\author{%
  Ryoichi Takase \quad
  Masaya Tsunokake \quad
  Yuta Tsuchiya \quad
  Shota Inuzuka \\
  Hitachi, Ltd. \\
}

\begin{document}
\maketitle

\begin{abstract}
Mathematical reasoning problems are among the most challenging, as they typically require an understanding of fundamental laws to solve. The laws are universal, but the derivation of the final answer changes depending on how a problem is approached. When training large language models (LLMs), learning the capability of generating such multiple solutions is essential to accelerate their use in mathematical education. To this end, we train LLMs using generative flow network (GFlowNet). Different from reward-maximizing reinforcement learning (RL), GFlowNet fine-tuning seeks to find diverse solutions by training the LLM whose distribution is proportional to a reward function. In numerical experiments, we evaluate GFlowNet fine-tuning and reward-maximizing RL in terms of accuracy and diversity. The results show that GFlowNet fine-tuning derives correct final answers from diverse intermediate reasoning steps, indicating the improvement of the capability of alternative solution generation.
\end{abstract}

\section{Introduction}
In mathematics education, it is recommended that teachers encourage students to seek alternative solutions rather than to find a single correct solution~\cite{stipek1998value,stipek2001teachers}. Considering alternative solutions helps students improve their ability to solve problems from different perspectives and creatively~\cite{lee2011effect,purwasih2019analysis}. To provide such learning environments for students, teachers need to understand alternative solutions. Recent advances in large language models (LLMs) have led to exploration of their use in mathematics education~\cite{kumar2023math,yue2024mathvc}. If the LLMs are used as teachers or used to support education, ideally, LLMs should have the same level of understanding of solutions as human teachers. Although many studies focus on improving accuracy~\cite{wang2022self,naik2023diversity,li2022making,lightman2023let,uesato2022solving}, to achieve better mathematics education using LLMs, we also need to address the capability of generating alternative solutions.

One way to train LLMs that can generate such solutions is generative flow network (GFlowNet;~\cite{bengio2021flow,bengio2023gflownet}).
Unlike reward-maximizing reinforcement learning (RL), GFlowNet trains a model whose probability distribution is proportional to a reward function.
The generated sequences are sampled from the distribution of the reward function, which allows us to obtain diverse, high-reward sequences~\cite{malkin2022trajectory,madan2023learning,hu2023amortizing,yu2024flow}.
In this study, we apply GFlowNet fine-tuning to mathematical reasoning tasks, which are among the most challenging since they typically require multi-step reasoning to solve.
The advantages of GFlowNet, which allows for the generation of diverse, high-reward sequences, would improve diversity of intermediate reasoning steps while maintaining the correctness of the final answer.
This means that GFlowNet has the potential to be suitable for alternative solution generation, but its capability is rarely discussed.

The objective of this study is to investigate the advantages of GFlowNet fine-tuning in mathematical reasoning tasks in terms of accuracy and diversity.
Among the solutions generated by LLMs, we are interested in correct ones that solve the problem by applying the mathematical laws from a different perspective.
These solutions derive the same and correct final answer from the problem, but the intermediate reasoning steps are less similar.
For clarity, we summarize what the following words refer to:
{
\begin{itemize}
\vspace{-2mm}
\item \textit{Solution}: Combination of intermediate reasoning steps and a final answer,
\item \textit{Distinct correct solutions}: Solutions where their intermediate reasoning steps can be distinguished and their final answers are correct for a given problem,
\item \textit{Diversity}: Less similarity of intermediate reasoning steps. In this study, we consider two intermediate reasoning steps to be distinct if their similarity is less than a threshold.
\vspace{-1mm}
\end{itemize}
}
Distinct correct solutions corresponds to a situation where the final answer is the same and correct, but the derivation is different (i.e., alternative solutions) in mathematics.
Our focus is on exploring the capability of such solution generation for GFlowNet and reward-maximizing RL.
The contribution of this study is to answer the following research questions:
\begin{itemize}
\vspace{-1mm}
\item Can GFlowNet generate diverse solutions in mathematical reasoning tasks?
\item What are the differences in accuracy between GFlowNet and reward-maximizing RL?
\vspace{-3mm}
\end{itemize}

\begin{figure}[t]
\centering
\includegraphics[width=1.0\textwidth]{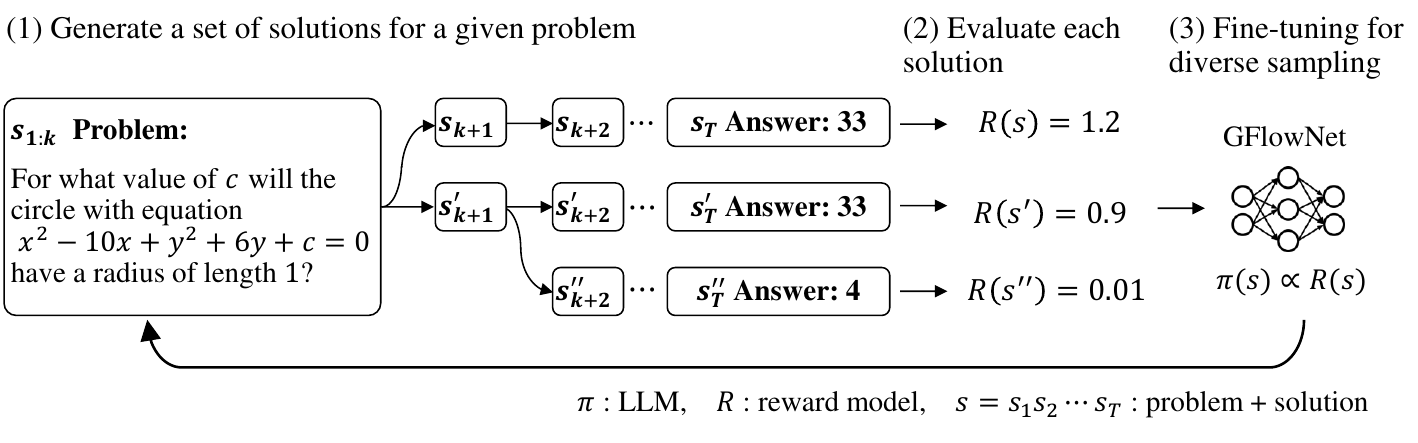}
\caption{GFlowNet fine-tuning consists of three steps: (1) multiple solutions are sampled for a given problem, (2) each solution is evaluated by a reward model, and (3) the parameters of an LLM are updated so that $\pi(s)\propto R(s)$. The solutions are sampled from the distribution of $R(s)$, which allows us to prevent the LLM from only generating high-reward answers.}
\label{fig:overview}
\vspace{-2mm}
\end{figure}

\section{Related Works}
\paragraph{Mathematical Reasoning.}
To enhance mathematical reasoning capabilities, prompting and decoding methods have been reported~\cite{wang2022self,naik2023diversity,li2022making}, improving accuracy by sampling multiple solutions and selecting a potentially correct one.
The use of evaluation models, especially process-level evaluations, has been shown to be effective in improving accuracy~\cite{lightman2023let,uesato2022solving}.

\paragraph{RLHF.}
After the success of training models to align human preferences~\cite{ouyang2022training}, reward maximizing RL such as proximal policy optimization (PPO)~\cite{schulman2017proximal} and direct preference optimization (DPO)~\cite{rafailov2024direct} has been applied to mathematical reasoning tasks~\cite{sun2024easy,luong2024reft}.
In RL, the evaluation models is used as the reward models, so the improvement of the evaluators and the connection with RL is one of the main research topics~\cite{wang2024math}.

\paragraph{GFlowNet.}
GFlowNet~\cite{bengio2021flow,bengio2023gflownet} has been proposed to obtain diverse and high quality samples, improving success rate in biological sequence design~\cite{jain2022biological,jain2023multi}.
GFlowNet has been used to fine-tune LLMs and has shown improved performance in story infilling and logical reasoning tasks~\cite{hu2023amortizing,yu2024flow}.

Our work continues these previous researches and explores the capability of alternative solution generation. 
The advantages of GFlowNet fine-tuning are investigated in terms of accuracy and diversity through experiments.

\section{Method}
\paragraph{Language Model.}
Let us consider an LLM $\pi_\theta$ with parameters $\theta$ is designed to generate a solution~$Y$ for a given problem~$X$.
The distribution over sequences of tokens is represented by an auto-regressive manner as follows.
\begin{align}
\pi_\theta(Y|X)=\prod_i \pi_\theta(y_i|X,y_{1:i-1}),
\end{align}
where $y_i$ is the $i$-th token in the solution and $y_{1:i-1}$ is tokens in the solution before $y_i$, which means that
the next token is sampled from the distribution conditioned before its token.

\paragraph{GFlowNet.}
To generate diverse solutions, we use the framework of GFlowNet.
Following~\cite{malkin2022trajectory,hu2023amortizing}, the notations of the problem $X$ and the solution $Y$ are respectively represented as $X=s_1s_2...s_{k}$ and $Y=s_{k+1}s_{k+2}...s_T$, where $s_i$ denotes the $i$-th token, $k$ denotes the token length of the problem, and~$T$ denotes the stop symbol of sequence generation.

Figure~\ref{fig:overview} shows the overview of GFlowNet fine-tuning.
For a given problem, the LLM first generate multiple solutions.
The solutions are then provided to a reward model that predicts their correctness as rewards.
Finally, the parameters of the LLM are updated by minimizing GFlowNet loss.
In this study, we use a modified version of the subtrajectory balance (SubTB) loss~\cite{madan2023learning} as described in~\cite{hu2023amortizing}.
The learning objective is given as follows:
\begin{align}
\mathcal{L}=\sum_{0\leq i<j\leq n}\biggl(\log \frac{R(s_{1:i}s_T)\prod_{k=i+1}^j\pi_\theta(s_k|s_{1:k-1})\pi_\theta(s_T|s_{1:j})}{R(s_{1:j}s_T)\pi_\theta(s_T|s_{1:i})} \biggr)^2,
\end{align}
where $R$ is the reward function evaluating the correctness of the solutions.
After learning converges, we obtain the LLM such that $\pi(s)\propto R(s)$, see~\cite{bengio2021flow,bengio2023gflownet,madan2023learning} for the details.

Different from reward-maximizing RL, our approach seeks to obtain high-quality, diverse solutions by training the LLM whose distribution is proportional to the reward function.
Specifically, the solutions are sampled from the distribution of the reward function, which allows us to prevent that only highest-reward solutions are generated.
The advantages of GFlowNet fine-tuning are discussed in the next section.

\section{Experiments}
\subsection{Setup}

\paragraph{Datasets.}
Our experiments are carried out on two widely used datasets for mathematical reasoning: GSM8K~\cite{cobbe2021training} and MATH~\cite{hendrycks2021measuring}.
GSM8K contains grade school math word problems that take between~2 and~8 steps to solve. 
MATH contains problems from mathematics competitions, including a full step-by-step solution for learning answer derivations.
For evaluation, we use the whole test set of GSM8K and use the subset MATH500 used in~\cite{lightman2023let}.
For the training of the LLMs and reward models, we use the following datasets derived from the above.
Regarding the training of the LLMs, we use MetaMATH~\cite{yu2023metamath} that is augmented from the original training sets to improve quality and diversity.
Regarding the training of the reward models, we use MathShephred~\cite{wang2024math} that contains step-level labels for learning whether current reasoning steps are correct.

\paragraph{Metrics.}
For the accuracy of the predicted final answer, the evaluation metrics used are the final answer accuracy of the greedy decoding and the best-of-K solution (i.e., pass@$k$).
Following~\cite{lightman2023let}, we round numerical values and parse expressions with \texttt{sympy}\footnote{Our evaluation code is based on \url{https://github.com/ZubinGou/math-evaluation-harness}.}.
For diversity of the predicted solutions, we compute the number of the distinct correct solutions.
Following~\cite{yang2024weak}, we consider two solutions distinct if their ROUGE-L similarity is less than~0.7.
Diversity analysis can be performed with a relatively small number of solutions~\cite{yang2024weak}, so we sample~$k=8$ solutions for each problem.

\paragraph{Baselines.}
We compare GFlowNet fine-tuning to the following reward maximizing methods: Rejection sampling fine-tuning (RFT)~\cite{yuan2023scaling}, Direct policy optimization (DPO)~\cite{rafailov2024direct}, and Proximal policy optimization (PPO)~\cite{schulman2017proximal}.
In this work, we use Llama3-8B base~\cite{dubey2024llama} as the base language model.
We use LoRA~\cite{hu2021lora} in all experiments.
For evaluation, we use the same 8-shot chain-of-thought prompt as in~\cite{wei2022chain} for GSM8K, and use the 4-shot problems available in~\cite{lewkowycz2022solving} for MATH.
The implementation details are described in appendix~\ref{sec:implementation}.

\subsection{Main Results}

\begin{figure}[t]
\begin{minipage}[b]{0.98\linewidth}
\centering
\end{minipage}\\
\begin{minipage}[b]{0.49\linewidth}
\centering
\includegraphics[keepaspectratio, scale=0.6]{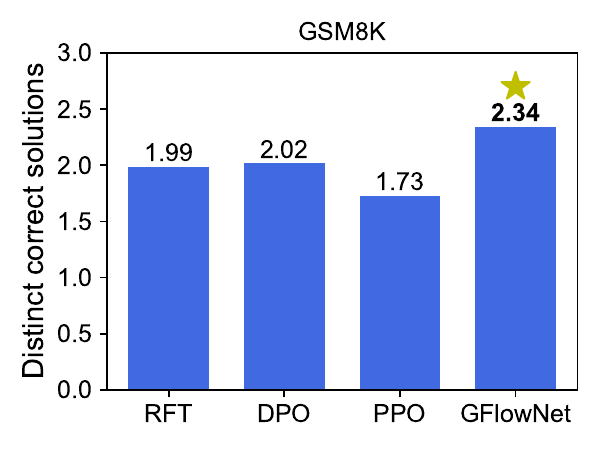}
\end{minipage}
\begin{minipage}[b]{0.49\linewidth}
\centering
\includegraphics[keepaspectratio, scale=0.6]{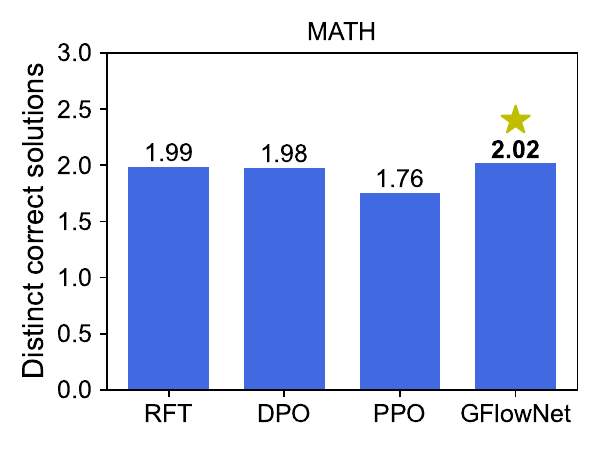}
\end{minipage}
\caption{The number of distinct correct solutions when sampling $k=8$ solutions. The star symbol indicates the method that has produced the highest number of distinct correct solutions.}
\label{fig:distinct_solution}
\end{figure}

\paragraph{GFlowNet derives correct answers from diverse intermediate reasoning steps.}
Figure~\ref{fig:distinct_solution} shows the number of distinct correct solutions when sampling $k=8$ solutions.
GFlowNet has produced the highest number of distinct correct solutions on both GSM8K and MATH, generating two or more alternative correct solutions.
Meanwhile, the reward-maximizing methods tend to generate similar correct solutions, e.g., PPO has the lowest diversity, with the number of distinct correct solutions of under 1.8.
Therefore, GFlowNet fine-tuning generates diverse intermediate reasoning steps but they converse to the same and correct final answer.

\paragraph{Similar accuracy between GFlowNet and reward-maximizing methods.}
Table\ref{tab:performance-comparison} shows the final answer accuracy under various decoding settings.
For reference, we also include the results of Llama3-8B base without training, labeled Few-shot\footnote{Since our focus is on accuracy comparison of GFlowNet and reward-maximizing methods and not building a SOTA model, we leave further improvement with full parameter fine-tuning for future work.}.
Consistent with related works, all fine-tuning based methods outperform Few-shot.
Furthermore, there is no significant difference in accuracy between GFlowNet and the reward maximizing methods.
These results indicate that GFlowNet fine-tuning has the advantage of improving diversity while maintaining accuracy in mathematical reasoning tasks.

{
\tabcolsep = 4pt

\begin{table}[t]
\centering
\small
\begin{tabular}{ll|ccc|cccccccc}\toprule
Task	&	& \multicolumn{3}{c|}{GSM8K}	&	\multicolumn{3}{c}{MATH} 	\\\midrule	
Model	&	Method	&	greedy	&	pass@4	&	pass@8	&	greedy	&	pass@4	&	pass@8	\\\midrule
\multirow{5}{*}{Llama3-8B}	&	Few-shot	&	0.446	&	0.694	&	0.797	&	0.188	&	0.342	&	0.458	\\\cmidrule{2-8}
	&	RFT	&	0.695	&	0.856	&	0.900	&	0.266	&	0.414	&	0.492	\\
	&	DPO	&	0.692	&	0.861	&	0.906	&	0.264	&	0.408	&	0.498	\\
	&	PPO	&	0.692	&	0.848	&	0.892	&	0.256	&	0.419	&	0.516	\\
	&	GFlowNet	&	0.700	&	0.859	&	0.905	&	0.224	&	0.399	&	0.500	\\\midrule
\end{tabular}
\caption{Final answer accuracy under various decoding settings on GSM8K and MATH.
There is no significant difference in accuracy between GFlowNet and reward-maximizing methods.
}
\label{tab:performance-comparison}
\end{table}
}

\section{Conclusion}
In this study, we have explored the advantages of GFlowNet fine-tuning in mathematical reasoning tasks. Different from reward-maximizing RL, GFlowNet trains the LLM whose distribution is proportional to the reward function, which allows us to obtain the LLMs that generate correct, diverse solutions. The results have shown that GFlowNet fine-tuning generates distinct derivations but they lead to the same and correct final answer.
To obtain more comprehensive results, our future work will include full parameter fine-tuning and evaluation on different tasks using other baseline models.

{
\small
\bibliographystyle{plain}
\bibliography{ms}
}

\newpage
\appendix

\section{Implementation Details}
\label{sec:implementation}
In this study, we use LoRA~\cite{hu2021lora} with the dimension of 128 for all training.
Regarding the reward models, the Llama3-8B is trained on MathShephred using token classification training, where the step-level-labels are used as the token labels.
Regarding the LLMs, we first perform SFT on MetaMATH with~1 epoch.
Our SFT and PPO implementation is based on DeepSpeed-Chat~\cite{yao2023deepspeed}.
For the sampling settings, we sample solutions with the temperature of 0.6 and the top-p of 0.9.
The hyperparameters are described as follows.

\paragraph{RFT}
For RFT training, by using the SFT model, we sample solutions for each problem and select the one with the best reward.
In this study, we sample $k=4$ solutions since a value of $k$ equal to or greater than 3 is sufficient~\cite{yuan2023scaling}.
We use 10K samples of MetaMATH.
The learning rate is set to 3e-6.

\paragraph{DPO}
For DPO training, we use the regularization coefficient of $\beta=0.01$ with the learning rate of~3e-7.
Regarding the training data, we use 10K samples of MetaMATH and sample 4 solutions for each problem using the SFT model.
The preference pairs is created with the solutions that have the highest and the lowest reward.

\paragraph{PPO}
For PPO training, we separate the parameters of actor and critic models.
The learning rate is set to 3e-6 for actor and 5e-6 for critic.
We use the KL penalty coefficient $\beta=0.1$, the discount factor $\gamma=1.0$, and the advantage estimation factor $\lambda=0.95$.
We use problems from MetaMATH as our training dataset.
During training, 12.8K samples are used to update the parameters.

\paragraph{GFlowNet}
For GFlowNet training, we use a replay buffer with the size of 1,000.
The learning rate is set to 3e-6.
During training, we also use SFT loss with the coefficient of 30.0 in accordance with~\cite{ouyang2022training,yao2023deepspeed}.
We sample 8 solutions for a problem with MetaMATH.
During training, 12.8K samples are used to update the parameters.

\end{document}